\documentclass{article}

\usepackage{PRIMEarxiv}

\usepackage[utf8]{inputenc} % allow utf-8 input
\usepackage[T1]{fontenc}    % use 8-bit T1 fonts
\usepackage{hyperref}       % hyperlinks
\usepackage{url}            % simple URL typesetting
\usepackage{booktabs}       % professional-quality tables
\usepackage{amsfonts}       % blackboard math symbols
\usepackage{nicefrac}       % compact symbols for 1/2, etc.
\usepackage{microtype}      % microtypography
\usepackage{lipsum}
\usepackage{fancyhdr}       % header
\usepackage{graphicx}       % graphics
\graphicspath{{media/}}     % organize your images and other figures under media/ folder
\usepackage{xcolor}

\usepackage{amsmath,amsfonts,amssymb}
\usepackage{graphicx}
\usepackage{multirow}
\usepackage{booktabs}
\usepackage{float}

\usepackage{amsthm}
\newtheorem{theorem}{Theorem}%[section]

\newtheorem{lemma}[theorem]{Lemma}

\title{Enhancing Digital Hologram Reconstruction Using Reverse-Attention Loss for Untrained Physics-Driven Deep Learning Models with Uncertain Distance}

\author{
  Xiwen Chen, Hao Wang, Abolfazl Razi\thanks{Corresponding author.} \\
  School of Computing \\
  Clemson University \\
  Clemson, SC, USA\\
   \And
  Zhao Zhang \\
  Department of Bioengineering \\
  Clemson University \\
  Clemson, SC, USA\\
  \And
  Zhenmin Li \\
  School of Electrical and Computer Engineering \\
  Georgia Institute of Technology \\
  Atlanta, GA, USA \\
  \And
  Huayu Li  \\
  Department of Electrical and Computer Engineering\\
  The University of Arizona \\
  Tucson, AZ, USA \\
  \And
  Tong Ye\\
  Department of Regenerative Medicine and Cell Biology\\
  Medical University of South Carolina\\
  Charleston, SC, USA
}

\begin{document}
\maketitle

\begin{abstract}
Untrained Physics-based Deep Learning (DL) methods for digital holography have gained significant attention due to their benefits, such as not requiring an annotated training dataset, and providing interpretability since utilizing the governing laws of hologram formation. However, they are sensitive to the hard-to-obtain precise object distance from the imaging plane, posing the \textit{Autofocusing} challenge. 
Conventional solutions involve reconstructing image stacks for different potential distances and applying focus metrics to select the best results, which apparently is computationally inefficient. In contrast, recently developed DL-based methods treat it as a supervised task, which again needs annotated data and lacks generalizability. 
To address this issue, we propose \textit{reverse-attention loss}, a weighted sum of losses for all possible candidates with learnable weights. This is a pioneering approach to addressing the Autofocusing challenge in untrained deep-learning methods. 
Both theoretical analysis and experiments demonstrate its superiority in efficiency and accuracy. Interestingly, our method presents a significant reconstruction performance over rival methods (i.e. alternating descent-like optimization, non-weighted loss integration, and random distance assignment) and even is almost equal to that achieved with a precisely known object distance. For example, the difference is less than 1dB in PSNR and 0.002 in SSIM for the target sample in our experiment.
\end{abstract}

% keywords can be removed

\section{INTRODUCTION}\label{sec:intro}  % \label{} allows 
Digital Holography (DH) is an emerging microscopic imaging technique with applications in Internet of Thing (IoT) security \cite{matoba2009optical,rajput2015photon, wang2022fast, 10216773}, measurement tasks in production \cite{fratz2021digital}, non-invasive various diagnoses \cite{rong2015terahertz,el2020discrimination}, etc. The basic idea of DH is to use an imaging sensor to capture a hologram of real micro-scale samples. Subsequently, the object information is reproduced via numerical reconstruction from the hologram. 
% Since the hologram preserves the depth information of small objects, numerical methods can be applied to reconstruct such spatial content \cite{}
Due to the powerful capacity of Deep Learning (DL) in recent years, the community has raised the trend to employ DL in their developed algorithms, which generally demonstrates the superiority over conventional physics-based calculation \cite{shimobaba2022deep}.

 \begin{figure}[htbp]
\centering\includegraphics[width=1\columnwidth]{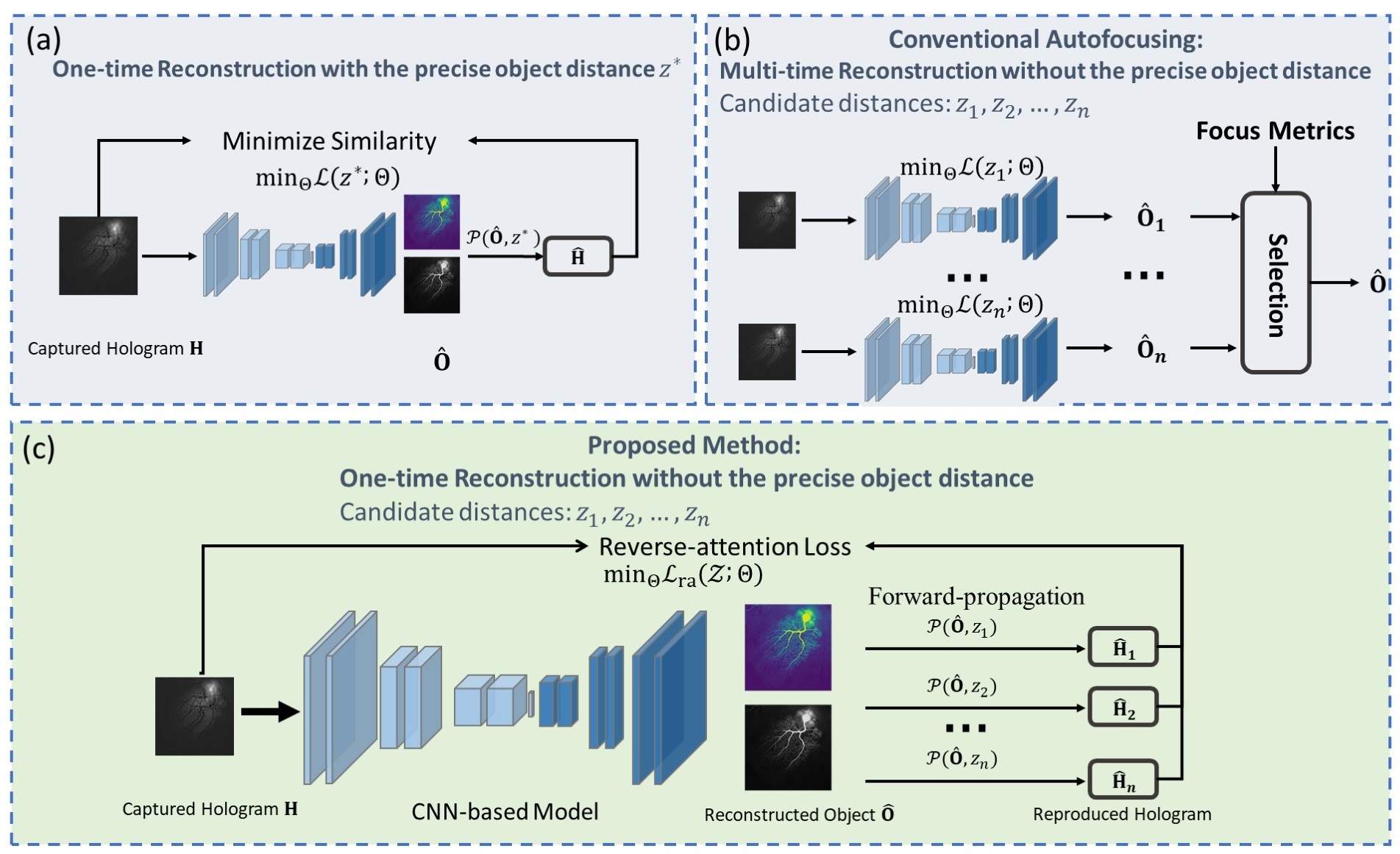}
\caption{Comparison of the untrained DL-based method for different scenarios: (a) the precise object distance is known, (b) the precise object distance is unknown and using conventional Autofocusing methods, and (c) the precise object distance is unknown and using our proposed method. The sample is from \cite{P1170}.}
\label{fig:main}
\end{figure}

\begin{figure}[ht]
\centering\includegraphics[width=1\columnwidth]{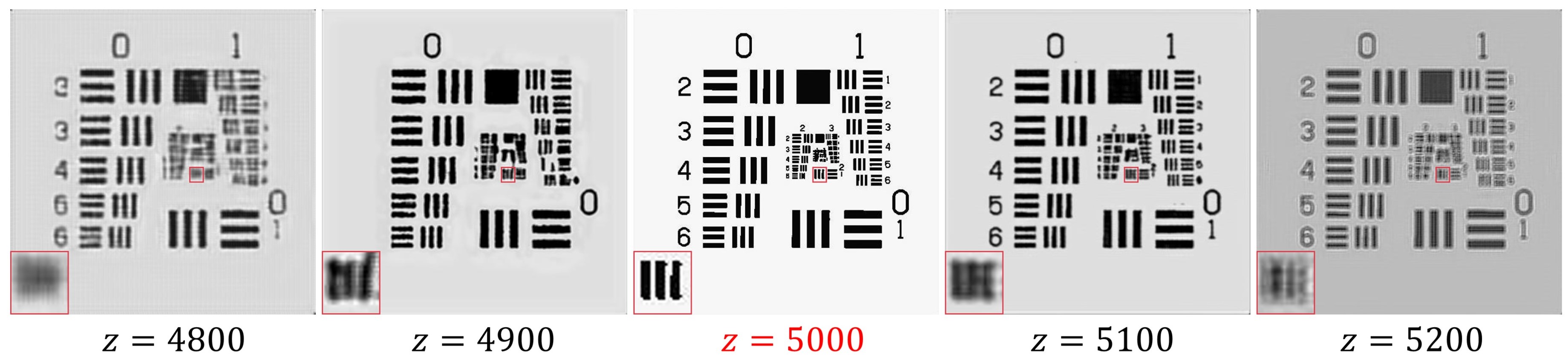}
\caption{The example of reconstruction by DeepDIH using different object distances. The actual distance is $5000\mu m$.}
\label{fig:example}
\end{figure}

In this work, we focus on Digital In-line Holography (DIH), known for its twin-image problem.
% which involves a known issue, termed twin-image removal.
To this end, several DL models are developed \cite{wang2018eholonet,horisaki2018deep,rivenson2018phase,zhang2020deep,wang2019net,ren2019end,chen2022fourier,wu2019bright} to reconstruct images based on end-to-end learning from multiple annotated hologram recordings.
% , which end-to-end learn the reconstruction process directly from annotated multiple hologram recordings.
When in the stage of inference, given the captured hologram, these models can automatically remove the twin image and reconstruct the object without feeding extra parameters. However, these methods usually suffer from the obvious drawback of reliance on relatively large datasets for training purposes. Since the interference fringes of holograms vary significantly depending on the holographic recording conditions and target objects, a hologram dataset for general purposes cannot be found. Therefore, it is necessary to create application-specific datasets, which require much effort \cite{shimobaba2022deep}. 
In some other applications, such as authenticating objects using nano-scaled visual tags, data sharing can be prohibited for security reasons \cite{chi2020consistency,10216773}. Hence, the untrained physics-driven models \cite{li2020deep,niknam2021holographic,bai2021dual,chen2023dh,manisha2023randomness,wang2024fast} have obtained impressive attention in more recent years since these models can reconstruct a twin image-free reproduced image using only a captured hologram without extra datasets and providing interpretability since utilizing the governing laws of hologram formation. As shown in Fig. \ref{fig:main}(a), the basic idea behind them is using a Convolutional Neural Network (CNN) as a function approximator to model the inverse process of the hologram generation, which maps the captured hologram ($z=d$) to the reconstructed complex-valued object ($z=0$). Subsequently, forward propagation is applied to the resulting complex-valued object to reproduce a hologram. The objective function is to minimize the distance between the synthesized hologram and the captured hologram to ensure the reliability of the learning inverse process approximation while the prior of CNN \cite{ulyanov2018deep} ensures the learned image close to a natural image, which automatically removes the twin-image artifact. However, these methods raise a new challenge that requires knowing the precise location of the object. An example is shown in Fig. \ref{fig:example}, the hologram is generated for the object at $z^*=5000\mu m$. Subsequently, we use DeepDIH \cite{li2020deep} to reconstruct it by using different object distances. 
The reconstruction can only be performed reasonably when the object is positioned at the correct distance (i.e. $z=5000 \mu m$).

This challenge, known as \textit{Autofocusing}, has also been studied in past years. The most conventional solution is reconstructing image stacks with all candidate depths and applying focus metrics (shown in Fig. \ref{fig:main}(b)), such as magnitude differential \cite{lyu2017fast}, entropy \cite{subbarao1998selecting,ilhan2014digital}, variance \cite{ren2015Autofocusing} and edge sparsity \cite{zhang2017edge}, to select the best results, which is inefficient and computational-costly. 
In contrast, recently developed DL-based methods treat it as a supervised task \cite{ren2018learning}, where the input is the captured hologram and the output is the predicted object position. Accordingly, this task can be formulated as a classification problem \cite{ren2018autofocusing,pitkaaho2019focus,cuenat2022convolutional} or regression problem \cite{ren2018learning,shimobaba2018convolutional,cuenat2022fast,montoya2023focusnet} when training the DL model. 
Similar to the supervised methods for DIH reconstruction, these methods again need an amount of annotated data and lack generalizability. Authors in \cite{tang2021single} predict the object distance by using an untrained network to simulate the forward process; however, this interesting attempt requires a known phase object, which is not applicable for reconstruction purposes. To the best of the authors' knowledge, there is no existing approach designed for untrained models used in DIH reconstruction to solve the Autofocusing issue.

To this end, we formulate it as a continuous-discrete optimization problem, which entails alternating optimization of DL model parameters (continuous) and object distance (discrete). Subsequently, we propose a \textit{reverse-attention loss}, a weighted sum of losses for all possible candidates with learnable weights, designed for untrained models that can learn the distance implicitly. As shown in Fig. \ref{fig:main}(b)(c), compared to the conventional methods that require multi-time reconstruction and then perform selection, our method allows the one-time reconstruction under an arbitrary number of candidate object distances. 
We theoretically analyze this loss and demonstrate its convergence rate aligns with the existing DL reconstruction method that uses a precise object distance. When integrating this plug-and-play technique into existing untrained models, its superiority is evident compared to rival methods (i.e. alternating descent-like optimization, non-weighted loss integration, random distance assignment) in our proof-of-concept experiments. Interestingly,  the reconstruction performance of the proposed framework is approximately equivalent to that achieved with a precisely known distance.

\section{Background Information}

\subsection{Reconstruction in DIH}
We first quickly recap the twin-image problem in the reconstruction of DIH. In DIH, a thin transparent object is presented as $\mathbf{O}(x,y;z=0)= \mathbf{R}(x,y;z=0)t(x,y)$, where $\mathbf{R}(x,y;z=0)$ is the reference wave (i.e., the incident wave if the object is not present) and $t(x,y)$ is the incurred perturbation term caused by the object. Subsequently, this object wave is forward-propagated to the detector plane $\mathbf{O}(x,y;z=d)$ at $z=d$ as,
\begin{align} \label{eq:prop1}
    \mathbf{O}(x,y;z=d) &= \mathbf{p}(\lambda,z=d)\circledast \mathbf{O}(x,y;z=0)  = \mathcal{F}^{-1}\{\mathbf{P}(\lambda,z=d)\cdot\mathcal{F}\{\mathbf{O}(x,y;z=0)\}\},
\end{align}
where $\lambda$ denotes the wavelength and $\circledast$ denotes the convolution operator. $\mathcal{F}\{\cdot\}$ and $\mathcal{F}^{-1}\{\cdot\}$ denote the direct and inverse Fourier transforms, respectively. $\mathbf{P}(\lambda,z)$ is the transfer function, defined as
\begin{align}  \label{eq:prop2}
  \mathbf{P}(\lambda,z) = \exp \left(\frac{2 \pi j z}{\lambda} \sqrt{1-\left(\lambda f_{x}\right)^{2}-\left(\lambda f_{y}\right)^{2}}\right), 
\end{align}
where $f_{x}$ and $f_{y}$ denote the spatial frequencies. Likewise, the reference wave can be propagated to $ \mathbf{R}(x,y;z=d)$. Then, the interference pattern due to the superposition of these two waves is captured by a detector (e.g., CCD or CMOS sensor) at $z=d$. Mathematically, the hologram is presented as, 
\begin{align}\label{eq:1}
    {\mathbf{H}}(x,y) & = |\mathbf{O}(x,y)+\mathbf{R}(x,y)|^2 
                    =|\mathbf{O}(x, y)|^{2}+|\mathbf{R}(x, y)|^{2}+\mathbf{R}^{*}(x, y) \mathbf{O}(x, y)+\mathbf{R}(x, y) \mathbf{O}^{*}(x, y),                    
\end{align}
where we assume that $z$ is the propagation direction and $x-y$ is the wavefront plane, and we omit the $z=d$ in this equation. 
$(\cdot)^*$ denotes complex conjugation. $\mathbf{R}^{*}(x, y) \mathbf{O}(x, y)+\mathbf{R}(x, y) \mathbf{O}^{*}(x, y)$ is the interference pattern.
In the reconstruction phase, the 3D information is retrieved by using the reference wave to illuminate the captured hologram,
\begin{align}\label{eq:2}
 \mathbf{R}(x,y){\mathbf{H}}(x,y)=|\mathbf{O}(x, y)|^{2}\mathbf{R}(x, y)+|\mathbf{R}(x, y)|^{2}\mathbf{R}(x, y)+|\mathbf{R}(x, y)|^{2} \mathbf{O}(x, y)+\mathbf{R}^{2}(x, y) \mathbf{O}^{*}(x, y),
\end{align}
 where the term $|\mathbf{R}(x, y)|^{2} \mathbf{O}(x, y)$ gives the distribution of the 3D object while the term $\mathbf{R}^{2}(x, y) \mathbf{O}^{*}(x, y)$ is a misleading term known as "twin-image".  Note that $\mathbf{R}^{*}(x, y) \mathbf{O}(x, y)$ and $\mathbf{R}(x, y) \mathbf{O}^{*}(x, y)$ are interchangeably consistent within the solution making the inverse problem under-determined.

\subsection{Untrained Physics-driven Method}\label{sec:relate2}
For clarification, the term \textit{Untrained} only means these methods do not require extra training data. For the sake of convenience, we use $\mathcal{P}(\cdot,z)$ and $\mathcal{P}^{-1}(\cdot,z)$ to denote the forward-propagation and back-propagation for a distance $z$, respectively. We use $\hat{\mathbf{O}}=G(\mathbf{H};\Theta)$ denotes a neural network with parameters $\Theta$ reconstruct the captured hologram $\mathbf{H}$ to $\hat{\mathbf{O}}$. Subsequently, a reproduced hologram by $\hat{\mathbf{O}}$ is denoted as $\hat{\mathbf{H}}=\mathcal{P}(\hat{\mathbf{O}},z)$. For simplicity, we describe the above to generate the reproduced hologram by $\hat{\mathbf{H}}:=f( \mathbf{H};z,\Theta)$. The objective function is,
\begin{align}\label{eq:loss0}
    \min_{\Theta}\mathcal{L}(z;\Theta) = d(\hat{\mathbf{H}},\mathbf{H}),
\end{align}
where $d(\cdot)$ denotes a metric space. Authors in \cite{li2020deep,niknam2021holographic,bai2021dual,manisha2023randomness,wang2024fast} use Mean-squared Error (MSE) as $d(\cdot)$, while \cite{chen2023dh} uses a discriminator to characterize the semantic differences. As discussed in Section \ref{sec:intro}, minimizing this distance guarantees the neural network learns a reliable inverse process, and Deep Image Prior (DIP) \cite{ulyanov2018deep} provides an implicit regularization to remove the twin image during the reconstruction. However, the performance of these methods heavily depends on knowing the accurate distance $z$, which is a problem that has not been solved yet.

\section{Problem Formulation}
Suppose an object at a distance $z^*$ and its captured hologram are denoted as $\mathbf{O}$ and $\mathbf{H}$, respectively. We adopt a similar setup of the conventional Autofocusing methods \cite{ilhan2014digital}, where we use a prior interval with a fixed step size as the candidate of object distances. Here, we denote these candidates as  $\mathcal{Z}=\{z_1,\cdots,z_n \}$. We assume $z^*\in \mathcal{Z}$. Our goal is to retrieve $\mathbf{O}$ from $\mathbf{H}$ by using these uncertain distances. Therefore, we have two variables $\theta$ (continuous) and $z_i$ (discrete) to optimize, resulting in a \textit{continuous-discrete} optimization problem.

\section{Methodology}

% For example, using DeepDIH \cite{li2020deep}

It is apparent that reconstructing by traversing all possible candidate distances is a super inefficient strategy for the Autofocusing problem since we should perform the reconstruction from scratch multiple times.  This is even worse for these physics-driven DL methods since DL methods sometimes have a larger memory and computation overhead. Therefore, we are looking for a method that only needs to perform reconstruction once and still can achieve a reasonable performance compared to the reconstruction with the known object distance. Intuitively, if a candidate distance yields a lower loss, it is more likely indicative of an accurate distance estimation. Hence, we can lead the model to pay more attention to this potential distance. To this end, we propose a novel loss. Since lower loss indicates greater attention, we name the proposed loss \textit{reverse-attention loss}.
\subsection{Reverse-attention Loss}
\begin{figure}[htbp]
\centering\includegraphics[width=1\columnwidth]{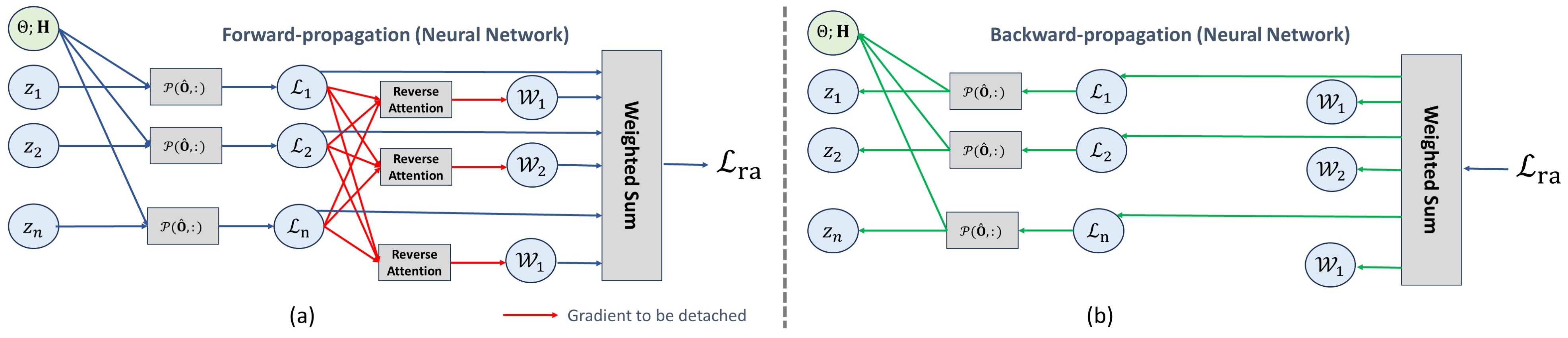}
\caption{The computational graph of the model $\Theta$ with the proposed loss. (a) Forward-propagation; (b) Backward-propagation. The \textcolor{red}{red} arrows denote the gradient of these weights, which will be detached in the backward-propagation. Note this propagation is for updating the neural network, which is different from the propagation of optical waves.}
\label{fig:computationgraph}
\end{figure}

 As discussed in Section \ref{sec:relate2}, suppose the reproduced hologram with distance $z_i$ by a DL model $\Theta$ is denoted as $\hat{\mathbf{H}}_i= f( \mathbf{H};z_i,\Theta)$. The loss for an untrained model is formulated by some measurement $d(\cdot)$ as $\mathcal{L}(z_i;\Theta)=d(\hat{\mathbf{H}}_i,\mathbf{H})$. To collectively learn the reconstruction for all candidates in one reconstruction, we propose the reverse-attention loss formulated as,
\begin{align}\label{eq:1}
    \min_{\Theta}\mathcal{L}_{ra}(\mathcal{Z};\Theta) = \sum_{i=1}^n \mathcal{W}(z_i) \mathcal{L}(z_i;\Theta), \quad\mathcal{W}(z_i;\Theta) = \frac{e^{1/\mathcal{L}(z_i;\Theta)}}{\sum_{i=1}^n e^{1/\mathcal{L}(z_i;\Theta)}}\quad(\textbf{Stop the gradient for }\mathcal{W}).
\end{align}
Apparently, $\sum_{i=1}^n\mathcal{W}(z_i;\Theta) = 1$ and $\mathcal{W}(z_i;\Theta)\geq 0$. To facilitate an easier flow of gradients, we apply the gradient detaching trick, which treats $\mathcal{W}(z_i)$ as a constant when performing the back-propagation of the network. In fact, we find this proposed loss has multiple benefits in the reconstruction problem, and using this detaching trick plays an important role that ensures the convexity of the proposed loss as discussed in the proof of THEOREM \ref{the:1}.
Fig. \ref{fig:computationgraph}(a)(b) show the forward-propagation and backward-propagation for the DL framework, respectively.
% When the loss for a $z_i$ is small, the attention $\mathcal{W}(z_i)$ to this distance is large. That is why we call it reverse attention. 

\subsection{Convergence Analysis}\label{sec:conv_ana}

% Now we adopt the common assumption in optimization \cite{}: i) 
% ii)
 Now, we adopt some assumptions commonly used in convergence analysis \cite{beck2009fast}. Suppose the function $\mathcal{L}(z_i;\Theta),\forall z_i\in\mathcal{Z}$ is convex and differentiable, and its gradient is Lipschitz continuous with constant $C\geq 0$, i.e. $\|\nabla \mathcal{L}(z_i;x)-\nabla\mathcal{L}(z_i;y)\|_2 \leq C\|x-y\|_2$ for any $x,y$.
We found the proposed loss has multiple good properties,
\begin{theorem}\label{the:0}
$\mathcal{L}_{ra}(\mathcal{Z};\Theta)$ has the same optimal $\Theta^*$ as $ \mathcal{L}(z^*;\Theta)$. i.e. when $ \mathcal{L}(z^*;\Theta^*)\rightarrow 0$, $ \mathcal{L}_{ra}(z^*;\Theta^*)\rightarrow 0$.
\end{theorem}
\begin{proof}
    It is obvious to obtain $\inf\mathcal{L}(z^*;\Theta)\rightarrow0$ and $\inf\mathcal{L}(z^*;\Theta)\leq \inf\mathcal{L}(z_i;\Theta), \forall z_i \in \mathcal{Z}$, since when the object distance is known, the optimal solution is achieved when the reproduced $\hat{\mathbf{H}}$ becomes exactly the same as $\mathbf{H}$ (i.e. $d(\hat{\mathbf{H}},\mathbf{H})=0$). Assume the optimal parameter is $\Theta^*$ and accordingly $\mathcal{L}(z^*;\Theta^*)\rightarrow 0$. Substitute it to Eq. (\ref{eq:1}), we can immediately imply that $\mathcal{W}(z^*;\Theta^*)= 1$, and subsequently $\mathcal{L}_{ra}(\mathcal{Z};\Theta^*) = \mathcal{L}(z^*;\Theta^*)$. The proof is completed. 
\end{proof}

\begin{theorem}\label{the:1}
   $\mathcal{L}_{ra}(\mathcal{Z};\Theta)$ has the same convergence rate as $ \mathcal{L}(z^*;\Theta)$ with respect to $\Theta$ if running gradient descent with a fixed step size $t\leq 1/C$.
\end{theorem}

  % \textit{Remark.} The convexity and Lipschitz continuity are commonly used in converge analysis, which is adopted in \cite{}.
\begin{lemma}\label{lemma:0}
    For any function $f(x):\mathbb{R}^m\rightarrow \mathbb{R}$ with the same assumption (i.e. differentiability, convexity and Lipschitz continuity) as $\mathcal{L}(z_i;\Theta)$,  if running gradient descent with a fixed step size $t\leq 1/C$ for $k$ iterations, it converges with rate $\mathcal{O}(1/k)$. The proof can be found in multiple materials, such as \cite{boyd2004convex,beck2009fast}.   
\end{lemma}
% Now we go back to Theorem \ref{the:0}.
\begin{proof}
It is apparent that at each moment (the time after we perform gradient detaching for $\mathcal{W}$), $ \mathcal{L}_{ra}$ is convex; this is because now $\mathcal{W}(z_i)$ is a constant and thus the summation of convex functions remains convex. Then, we evaluate the Lipschitz continuity of $\mathcal{L}_{ra}(\mathcal{Z};\Theta)$ for any $x,y$,
    \begin{align}
        &\|\nabla \mathcal{L}_{ra}(\mathcal{Z};x)-\nabla\mathcal{L}_{ra}(\mathcal{Z};y)\|_2 \\ \nonumber
        =&\|  \sum_{i=1}^n \mathcal{W}(z_i) \nabla\mathcal{L}(z_i;x)-\sum_{i=1}^n \mathcal{W}(z_i) \nabla\mathcal{L}(z_i;y)              \|_2 \\ \nonumber
        =&\| \sum_{i=1}^n \Large(\mathcal{W}(z_i)\left(\nabla\mathcal{L}(z_i;x)-\nabla\mathcal{L}(z_i;y) \right)\Large)     \|_2 \\ \nonumber
        \overset{(a)}{\leq} &\sum_{i=1}^n\mathcal{W}(z_i) \|\nabla\mathcal{L}(z_i;x)-\nabla\mathcal{L}(z_i;y) \|_2 \\ \nonumber
        \overset{(b)}{\leq} & \sum_{i=1}^n\mathcal{W}(z_i) C\|x-y\|_2 \\ \nonumber
        = &C\|x-y\|_2,
    \end{align}
    where (a) applies the triangle inequality of norm space and (b) applies the Lipschitz continuity of $\mathcal{L}(z_i;\Theta)$. This equation shows at each moment, $\mathcal{L}_{ra}(\mathcal{Z};\Theta)$ has the same property of $\mathcal{L}(z_i;\Theta)$. Then, after applying LEMMA \ref{lemma:0}, we can immediately confirm they have the same convergence rate at each moment for updating neural network. The proof is completed.
\end{proof}

\section{Experiment}
\subsection{Synthetic Analysis of Convergence}
\begin{figure}[hb]
\centering\includegraphics[width=1\columnwidth]{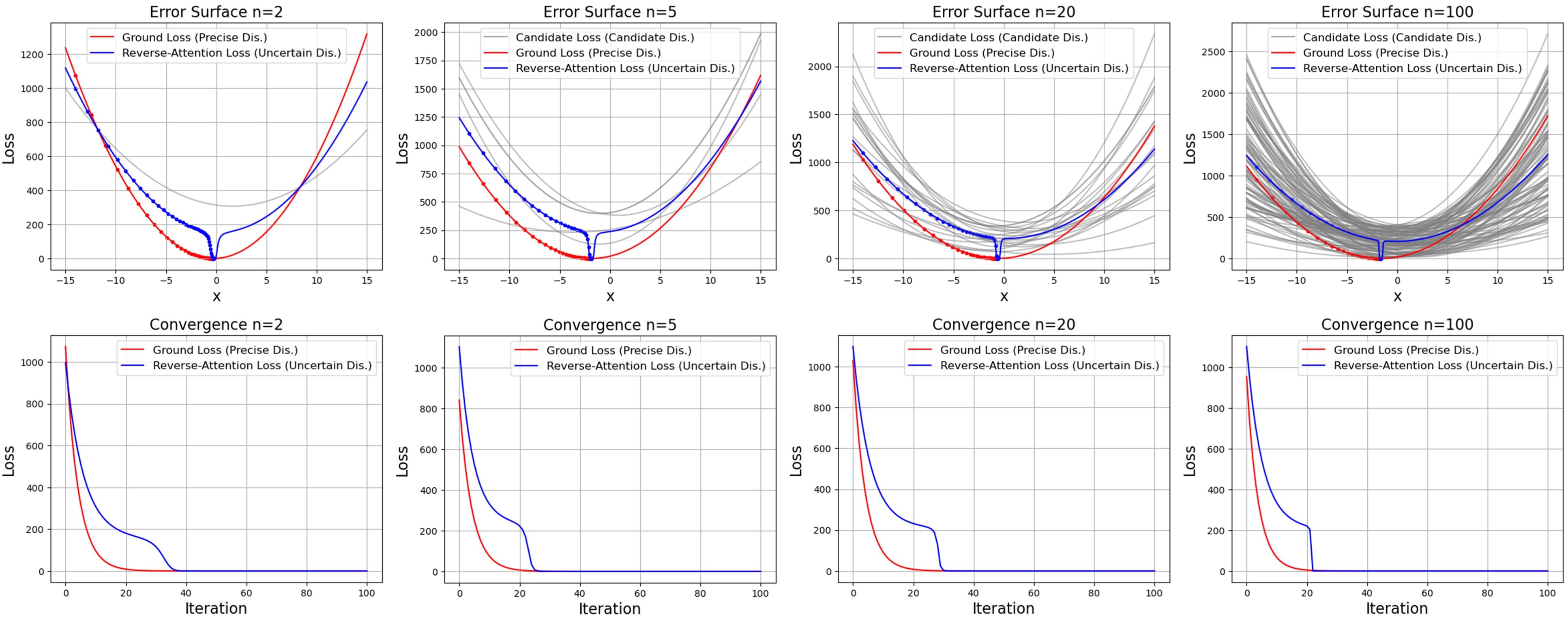}
\caption{The synthetic noisy quadratic analysis of convergence for $n= 2, 5,20,100$. \textbf{First Row:} The error surface of candidate losses (gray), the ground loss (red) (i.e. with the precise distance), and reverse-attention loss (blue). The dot point presents the update step with a known object distance (red) or without a known object distance (blue). \textbf{First Row:} The convergence rate of the ground loss and reverse-attention loss.}
\label{fig:cov_num}
\end{figure}

We first use synthetic noisy quadratic analysis \cite{zhang2019lookahead} to evaluate the convergence of the reverse-attention loss. We generate a number of quadratic functions $f_i(x)=(a_i  x - b_i)^2 + c_i$, where $a_i\sim \mathcal{U}(1,3)$, $b_i\sim \mathcal{U}(-5,5)$, and $c_i\sim \mathcal{U}(0,400)$. Here $f_i(x)$ can be considered the loss function with a candidate distance $z_i$, and $x$ is the optimization variable. The $f_i$ with a precise $z_i$ is generated by enforcing $c_i=0$, which ensures its minimum to zero. This setup is consistent with the scenario of DIH reconstruction discussed in Section \ref{sec:relate2} and Section \ref{sec:conv_ana}, which indicates that we can only have a perfect reconstruction with a precise distance (loss to 0). After generating all candidates, accordingly, the reverse-attention loss is generated by Eq. (\ref{eq:1}). 
As shown in Fig. \ref{fig:cov_num}, we perform 4 experiments with $n= 2, 5,20,100$, denoting the number of candidates. The first row presents the error surface of different candidate losses, the ground loss (i.e. with precise distance), and reverse-attention loss. The second row presents the convergence rate of the ground loss and reverse-attention loss. These results illustrate that reverse-attention loss has the same optima as the ground loss, and their convergence rate is similar, even when the number of candidates increases.

\subsection{Experiments on Samples}
\noindent\textbf{Setup.} As a proof-of-concept, we conduct the experiments by using the simulated holograms from three samples, including the \textbf{1951 USAF Resolution Target}, \textbf{Cell} \cite{zhang2018twin}, and \textbf{Dendrite Tag} \cite{chi2020consistency, wang2022fast}. 
The holograms are generated for the objects at $z^*=5000\mu m$ with 2.0 ${\mu m}$ pixel size and $532 nm$ wavelength. The sample images are resized to 500 $\times$ 500 pixels to align with the previous benchmarks \cite{zhang2018twin, li2020deep, chen2023dh}.
Our reconstruction is from $4500\mu m$ to $5500\mu m$ with a step size of $100\mu m$. 
This results in a set of candidate distances $\mathcal{Z}=\{4500um,4600um,4700um,\cdots,5500um\}$. 
We choose DeepDIH \cite{li2020deep} as the baseline model in this work since DeepDIH is one of the most representative untrained physics-driven DL methods for DH  reconstruction, which uses an auto-encoder structure. It is noteworthy that this architecture has been employed in the several following works \cite{niknam2021holographic,bai2021dual,manisha2023randomness,wang2024fast}.

% We use the Adam optimizer \cite{kingma2014adam} with a fixed learning rate of 1e-3.
\noindent\textbf{Baselines.}  We include multiple methods for comparison: \textbf{Know Precise Dis.}: the reconstruction by using the precise object distance. This can be seen as the upper bound for all Autofocusing methods. \textbf{Random Distance Assignment}: the reconstruction by using a random candidate distance. We compute the expected value of reconstruction performance for all possible candidates. \textbf{Include Correct Dis.} indicates we include the reconstruction by using the correct object distance in this computation, while \textbf{Exclude Correct Dis.} indicates we do not include it. The aforementioned one is used to demonstrate the difference in performance by using incorrect distances. We also consider two methods that optimize the two variables $\Theta$ and $z$ alternatively.
\textbf{Alternating Descent}: In each epoch, we compute the loss for different candidates $\mathcal{L}(z_1;\Theta),\mathcal{L}(z_2;\Theta),\cdots$. Then, choose the smallest one as the loss (i.e. $\mathcal{L}(\Theta)=\min_i \mathcal{L}(z_i;\Theta)$) to perform backward for updating the neural network. It is named \textit{Alternating} because we first optimize the parameter $z$ and then optimize the parameter $\Theta$. 
\textbf{Non-weighted loss integration}: This method simply takes the average of the losses by different candidates, which is presented as $\mathcal{L}(\Theta)=\frac{1}{n}\sum_i \mathcal{L}(z_i;\Theta)$.  All baselines are training based on DeepDIH.

\noindent\textbf{Evaluation Metrics.} To evaluate the reconstruction performance, we employ the Peak Signal-to-Noise Ratio (PSNR) and Structural Similarity Index Measure (SSIM), which are two widely used metrics for assessing the quality of images. We also test the runtime for different losses (i.e. $\min_{\Theta}\mathcal{L}(z;\Theta)$ and $\mathcal{L}_{ra}(\mathcal{Z};\Theta)$).

\noindent\textbf{Implementation Detail.} We use the Adam optimizer \cite{kingma2014adam} with a fixed learning rate of 1e-3. We use a PC with an Nvidia RTX 3080 graphic card and an Intel(R) Core(TM) i7-10870H CPU @ 2.20GHz, 2208 Mhz, 8 Core(s) for the runtime test.

\begin{table}[]
\centering
\caption{Comparison of the reconstruction performance by different Autofocsing Strategies.}
\label{tab:per}
\resizebox{\textwidth}{!}{%
\begin{tabular}{cc|ccccc|c} \toprule
\multirow{2}{*}{\textbf{Samples}} & \multirow{2}{*}{\textbf{Metrics}} & \multicolumn{5}{c|}{\textbf{Not Know Precise Object Dis.}} & \textbf{Know Precise Dis.} \\ \cline{3-8}
 &  & \textbf{\begin{tabular}[c]{@{}c@{}}Random   Distance\\      (Exclude Correct Dis.)\end{tabular}} & \textbf{\begin{tabular}[c]{@{}c@{}}Random Distance\\      (Include Correct Dis.)\end{tabular}} & \textbf{Non-weighted Integ.} & \textbf{Alternating} & \textbf{Ours} & \textbf{Upper bound} \\ \midrule 
\multirow{2}{*}{\textbf{Target}} & PNSR & 12.788 & 14.527 & 15.150 & 13.036 & \textbf{30.956} & 31.916 \\
 & SSIM & 0.753 & 0.774 & 0.803 & 0.815 & \textbf{0.973} & 0.975 \\ \midrule
\multirow{2}{*}{\textbf{Cell}} & PNSR & 18.848 & 20.415 & 17.763 & 19.469 & \textbf{32.522} & 34.248 \\
 & SSIM & 0.679 & 0.707 & 0.719 & 0.671 & \textbf{0.982} & 0.986 \\ \midrule
\multirow{2}{*}{\textbf{Dendrite Tag}} & PNSR & 22.368 & 23.379 & 22.954 & 21.437 & \textbf{28.762} & 33.491 \\
 & SSIM & 0.735 & 0.756 & 0.772 & 0.688 & \textbf{0.955} & 0.971 \\ \bottomrule
\end{tabular}%
}
\end{table}

\begin{figure}[h]
\centering\includegraphics[width=0.8\columnwidth]{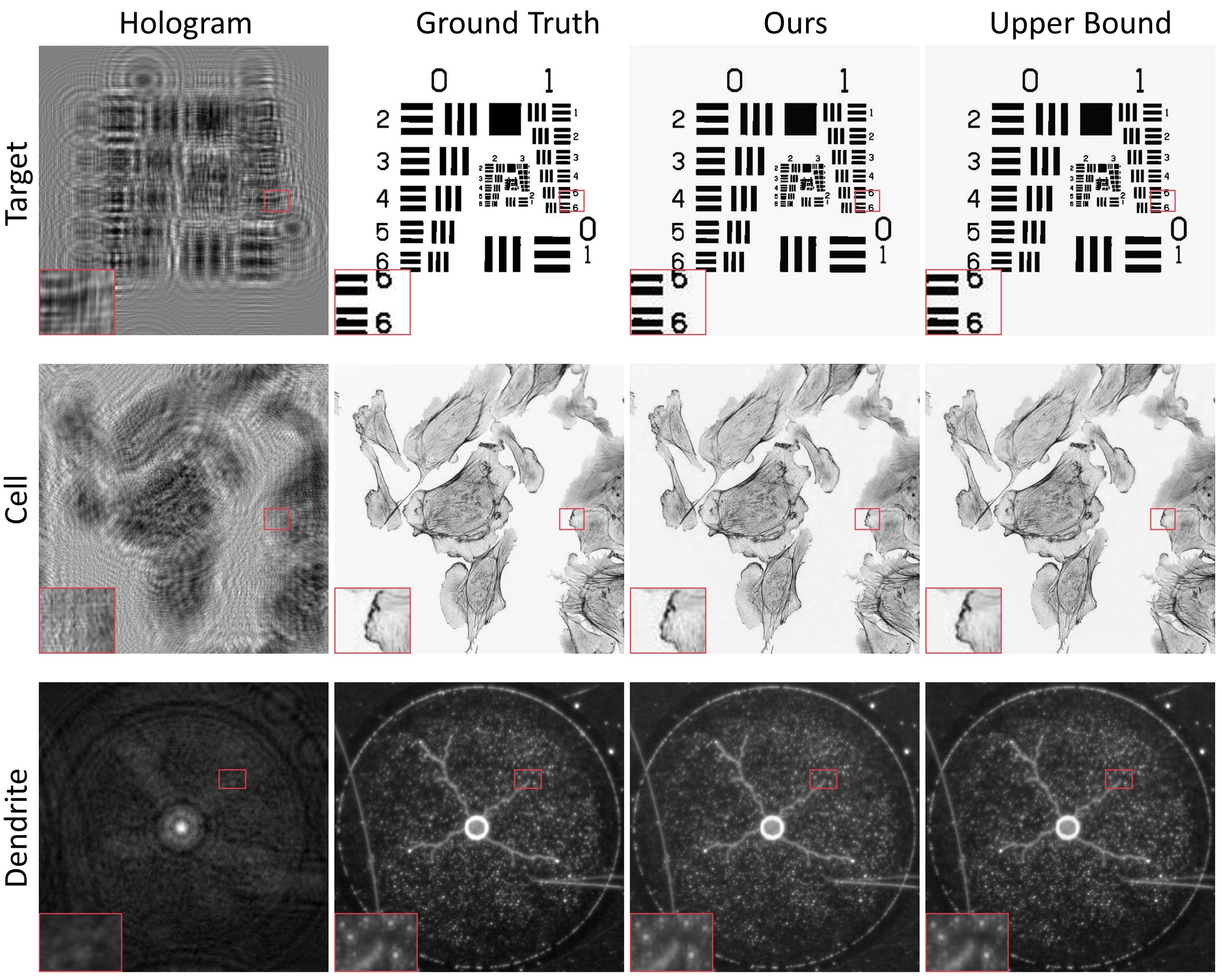}
\caption{Learning with uncertain accurate distance. \textbf{First Row:} The ground truth; \textbf{Second Row:} The hologram; \textbf{Third Row:} The reconstruction with our reverse-attention loss; and \textbf{Fourth Row:} Upper bound of the Autofocusing. The reconstruction uses the precise object distance.}
\label{fig:result1}
\end{figure}

\noindent\textbf{Results.} The reconstruction results are shown in Table \ref{tab:per} and Fig. \ref{fig:result1}. It is observed that our proposed methods can obtain substantial gains compared with other methods that reconstruct under uncertain distances. For example, we obtain 15dB, 12dB, and 5dB PNSR gain over the second-best method for different samples, respectively. More importantly, we found that using our proposed method, the reconstruction can even be very approximate to the reconstruction by using the precise object distance (upper bound in Table \ref{tab:per}). Specifically, the differences are 1dB, 2dB and 5dB in PSNR and 0.002, 0.004, and 0.02 in SSIM for three samples, respectively. This is also evident by the visualization shown in Fig. \ref{fig:result1}, where we found both reconstructions by our method and the upper bound can achieve similar excellent performances. The results of runtime are shown in Table \ref{tab:time}, which demonstrates reconstruction by our proposed loss only introduces reasonable overhead ($\sim$ 8 minutes) compared to one-time reconstruction by using original loss (i.e. Eq. (\ref{eq:loss0})). That means if using the conventional Autofocusing methods as discussed in Section \ref{sec:intro}, we should take $n\times 7$ minutes for reconstruction while our method only takes 15 minutes, which is much more efficient.

\begin{table}[htbp]
\centering
\caption{Compassion of runtime for one-time reconstruction by using different loss.}
\label{tab:time}
\resizebox{0.5\textwidth}{!}{%
\begin{tabular}{c|c|c} \toprule
\textbf{Loss} &  \textbf{\begin{tabular}[c]{@{}c@{}}Reconstruction Time\\      (per 5000 epochs)\end{tabular}} &  \textbf{\begin{tabular}[c]{@{}c@{}}Reconstruction Time\\     with Autofocusing\end{tabular}} \\ \midrule
$\min_{\Theta}\mathcal{L}(z;\Theta)$ & $\sim$7 mins &  $\sim 11\times 7$ mins \\
$\mathcal{L}_{ra}(\mathcal{Z};\Theta)$ & $\sim$ 15 mins & \textbf{$\sim$ 15 mins}\\ \bottomrule
\end{tabular}%
}
\end{table}

\section{Conclusion}
% In this work, we propose a reverse-attention loss to solve the Autofocusing challenge for untrained DL methods. W
In this work, we propose reverse-attention loss as a novel approach to address the Autofocusing challenge in untrained deep learning methods for DIH reconstruction. We give the theoretical convergence analysis and conduct proof-of-concept experiments to illustrate its superiority. This innovative solution demonstrates the potential to enhance the effectiveness of deep learning models in dealing with uncertain distances, offering an impressive reconstruction performance that approximates the reconstruction by using a precise object distance.

\section*{Acknowledgments}
This work is supported in part by the USDA AFRI program under grant \#2020-67017-33078 and the National Science Foundation EPSCoR Program under NSF Award \#OIA-2242812.
% \newpage
% References
\bibliography{report} % bibliography data in report.bib
\bibliographystyle{spiebib} % makes bibtex use spiebib.bst
\end{document}